\documentclass{article}
\usepackage{booktabs}
\usepackage{graphicx}
\usepackage{microtype}
\usepackage{xurl}
\usepackage[letterpaper, margin=1in]{geometry}
\usepackage{float}
\usepackage[font=small,labelfont=bf,skip=2pt]{caption}

\begin{document}

\begin{center}
{\Large In-context learning enables continental-scale subsurface temperature prediction from sparse local observations\par}

\vspace{1em} 

{\small
\parbox{\textwidth}{\centering
Daniel O'Malley$^{*}$, Christopher W. Johnson, Javier E. Santos, Pablo Lara,
Sandro Malusà, Bharat Srikishan, John Kath, Arnab Mazumder, Mohamed Mehana,
David Coblentz, Nathan DeBardeleben, Earl Lawrence, and Hari Viswanathan\\
{\footnotesize Los Alamos National Laboratory}, $^{*}$\texttt{omalled@lanl.gov}
}\par}

\vspace{0em} 
\end{center}

\begin{abstract}
    Continental-scale knowledge of subsurface temperature is limited by the cost and sparsity of borehole measurements, but such information is essential for geothermal resource assessment and for understanding heat transport in the shallow crust.
    The thermal field reflects the interaction between lithology, crustal structure, radiogenic heat production, and advective fluid flow, sometimes producing sharp anomalies that are smoothed by conventional interpolation or difficult to capture with physical models.
    Here we introduce In-Context Earth, a transformer-based model that uses sparse local borehole observations as geological context to predict continuous temperature-at-depth fields with calibrated uncertainty.
    In the contiguous United States, the model achieves a mean absolute error of 4.7 $^\circ$C, outperforming the physics-informed Stanford Thermal Model, a model based on AlphaEarth embeddings, the multimodal Transparent Earth model, and universal kriging, while resolving sharper thermal gradients in geothermal provinces.
    Its uncertainty estimates are well calibrated, with a Kolmogorov–Smirnov statistic of 2.5\%.
    Without finetuning, the model adapts to Alberta, Australia, and the United Kingdom (UK) using only 20 local observations at inference time, maintaining high accuracy in geologically distinct test regions with a mean absolute error of 2.2 $^\circ$C in Alberta, 6.2 $^\circ$C in Australia, and 5.4 $^\circ$C in the UK.
    Interpretability analyses show that the model learns internal representations of subsurface properties it never observes during training, including seismic velocities, geochemistry, and crustal structure, and uses these representations in physically consistent ways.
    More broadly, this work shows that in-context learning can use sparse borehole observations for continental-scale subsurface characterization, without requiring dense measurements or region-specific retraining.
\end{abstract}
\vspace{-1em}

\begin{figure}[H]
    \centering
    \includegraphics[width=1\linewidth]{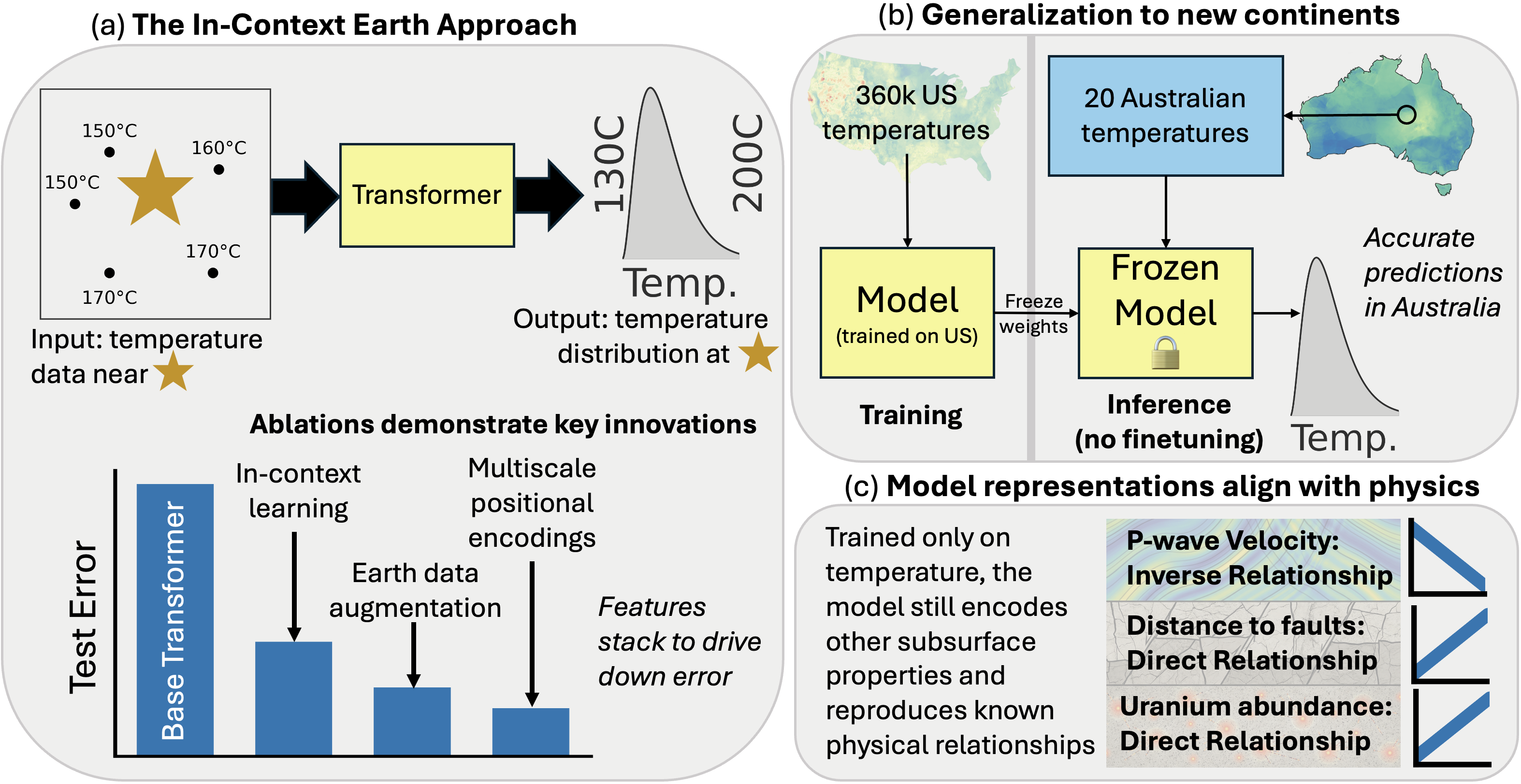}
    \caption{(a) The In-Context Earth approach uses transformers to make validated, uncertainty-aware predictions of geothermal temperature. Several innovations enable strong performance, including in-context learning, Earth-tailored data augmentation, and multiscale positional encodings. (b) Using training data from the US, the model shows good performance in other regions including Australia, Canada, and the UK. (c) Linear probes and representation interventions show that training with only temperature data still provides an approximate world representation that includes features like seismic velocities, faults, and geochemistry. 
    }
    \label{fig:workflow}
\end{figure}

\section*{Introduction}
Many subsurface datasets are sparse, and obtaining accurate and reliable information away from sites with direct measurements is a fundamental challenge in the geosciences.
Subsurface temperature is an excellent example since it is sparsely observed but has practical importance for defining the viability of geothermal resources \cite{blackwell2006assessment, tester2007future}.
Hence, accurate characterization of the shallow crust thermal regime is necessary for sustained geothermal energy production.
Despite the importance of subsurface temperature mapping, information about the shallow crust thermal regime is not well known. 
In contrast to densely sampled remote sensing measurements (e.g., satellite data), the cost of drilling and logging boreholes results in a significant undersampling of continental-scale thermal profiles relative to the spatial resolution of areal heat maps. 
Instead, maps are inferred from the available sparse, unevenly distributed measurements that are prohibitively expensive to collect.
Recent efforts have sought to integrate these sparse data into three-dimensional models to provide an image of the continental-scale crustal thermal structure \cite{blackwell2006assessment, blackwell2011temperature, boyd2019temperature, aljubran2024thermal}, yet data sparsity remains a fundamental limitation and increasing measurement density through additional drilling is not economically feasible.
Consequently, our understanding of thermal resources remains restricted to well-characterized sites, with many areas that may have high potential for geothermal energy development remaining poorly constrained.

Inferring shallow-crust thermal profiles with a deep learning model using sparse geophysical observations can provide a new representation of the interactions of lithology, crustal structure, fluid flow, and geochemistry in the subsurface and allow continental-scale temperature-at-depth maps (Figure \ref{fig:workflow}).
The subsurface thermal regime is not a simple conductive gradient and instead reflects many latent geological controls.
It is a complex pattern influenced by tectonic history and advective fluid flow, creating anomalies that complicate traditional interpretations \cite{kukkonen1994simulation}.
The complexity of the shallow crustal thermal profile and the sparsity of borehole measurements increases the difficulty of capturing the spatial heterogeneity of the crustal heat field and produces incomplete knowledge of the continental-scale thermal structure.
Landmark assessments have long advocated for geothermal development based on national temperature-at-depth maps \cite{tester2007future,williams2008assessment} and higher-fidelity regional maps provide valuable knowledge about high-temperature geothermal resources in the shallow crust (depths to about 3 km) that increase the estimated resource potential and improve the economics of early site development \cite{augustine2019geovision}.
Beyond energy resources, subsurface temperature also relates to the maturation of sedimentary basins, the integrity of carbon sequestration sites \cite{vilarrasa2017thermal}, and the rheological strength of the crust that controls deformation, faulting, and localization of tectonic strain \cite{brace1980limits,burov2006long}.
Successfully mapping the temperature would ideally require more than interpolation.
An internal model that combines subsurface properties and their role in transporting heat must be established, but is currently lacking.

Traditional geostatistical methods, such as universal kriging, are useful tools to extract knowledge in regions of sparse data.
This class of statistical methods relies on the assumption of local correlation structures to perform spatial interpolation using a variogram and a best linear unbiased estimator for spatially correlated data \cite{matheron1963principles}.
Since kriging enforces a variogram-governed smoothness through these simplified two-point correlation assumptions, these models can over-smooth the inferred field and not capture sharp gradients and localized anomalies that are observed in measurements.
Other spatial interpolation frameworks have been explored to handle irregular data, such as radial basis functions and thin-plate spline interpolation.
However, radial basis functions require careful parameter tuning to maintain accuracy \cite{fasshauer2007choosing}, while thin-plate spline interpolation risks erasing localized anomalies, and both tend to produce over-smoothed representations \cite{duchamp2003spline}.
Similarly, inverse distance weighting variants are simple heuristic smoothers that impose distance-only weighting but do not encode geological structure (e.g., faults, basin boundaries, advective pathways), systematically attenuate sharp gradients, and offer no principled uncertainty quantification \cite{li2011review}.
By treating geospatial estimation as a series of independent local problems, these approaches often fail to capture the broader geological context that dictates thermal regimes across continental scales.
Consequently, these methods struggle to resolve sharp thermal gradients or identify high-frequency information needed for precise geothermal resource characterization.
These limitations motivate deep learning approaches applied to subsurface temperature mapping that can learn many-point correlations, encode geologic structure, and translate this into accurate temperature predictions with uncertainty quantification.

This work focuses on a specific gap in geosciences -- a reliable and powerful in-context learning framework that uses sparse local observations \cite{dong2024survey}.
Emerging deep learning architectures offer a path forward by providing tools to discover complex patterns in the subsurface that are hiding in plain sight, but are not revealed by traditional geostatistical or physical models \cite{reichstein2019deep}.
In particular, transformer architectures based on self-attention \cite{vaswani2017attention} are powerful for learning long-range dependencies and in-context conditioning that enable large performance gains across domains, including natural language processing \cite{devlin2019bert,brown2020language}, computer vision \cite{dosovitskiy2021image}, and Earth system forecasting \cite{bodnar2025foundation}.
A major leap from GPT-2 to GPT-3 was few-shot in-context learning that allows the model to learn from a few examples after pretraining \cite{brown2020language}.
An example would be asking ChatGPT for a gift recommendation.
The response will be substantially improved if the user provides in-context information describing the age, interests, and personality characteristics of the individual.
Even though that specific information was not available during model pretraining, the in-context relations of the inputs will guide the model to improve the response.
Applying this same in-context approach with a deep learning modeling framework trained using existing sparse measurements is central to discovering the patterns that control subsurface heat flow.
Here, we show that a global, contextual, transformer-based learning approach produces high-resolution temperature-at-depth maps with calibrated uncertainty while learning an internal representation of the subsurface that retains anomalous hot spots and sharp lateral gradients.

\section*{Results}
\subsection*{In-Context Earth subsurface temperature model}
In-Context Earth is an encoder-only transformer designed to input a variable number of context tokens from multi-depth borehole measurements sampled from the surrounding geological region to inform the prediction at a specific target location (see Methods section and Supplement for details).
A multiscale spatial encoding mechanism is applied to represent the spatial relationships in the temperature distributions.
This design allows In-Context Earth to treat spatial inference of the temperature field as an in-context learning task.
The temperature is treated as a classification task rather than a regression task, which enabled robust uncertainty quantification.
We implement a conditioning mechanism where target temperatures are binned into $K$ discrete intervals (see Methods section) and task the model to predict the probability distribution across these bins.
This allows the model to represent non-Gaussian and other non-parametric uncertainty structures that are common in geologically complex regions.
Viewed through a language-model lens, the $K$ bins act as a small vocabulary and the transformer is trained to predict the masked temperature token conditioned on the context sequence.

The model is trained using a comprehensive compilation of borehole temperature measurements from the contiguous United States (US) \cite{aljubran2024thermal,aljubran2024stanford}.
The data contain well location coordinates, depths, and bottomhole temperatures. 
The input to the transformer consists of a sequence of tokens, where each token is a concatenated vector of the spatially encoded features ($X_i$) and its corresponding binned temperature ($y_i$).
To make in-context learning transferable to locations beyond the training data, we leveraged a data augmentation scheme that masks the latitude and longitude from the model, only providing the relative positions of data points.
This prevents the model from memorizing the geocoordinates of the temperature map and instead learning the in-context representation of the positionally encoded sparse measurements.
Table \ref{tab:ablation-training-details} in the supplementary information contains additional training details and hyperparameters used for our ablation studies.

We benchmark In-Context Earth temperature predictions against the Stanford Thermal Model (a physics-informed machine learning model) \cite{aljubran2024thermal}, a model that uses AlphaEarth embeddings \cite{nakata2026subsurface, brown2025alphaearth}, Transparent Earth (a multimodal data-driven approach) \cite{mazumder2025transparent}, and universal kriging (traditional geostatistical baseline) -- see Table \ref{tab:mae_us}.
All models were trained on data from the US and the results demonstrate the performance of each on in-distribution tasks.
We use mean absolute error (MAE) as our error metric for consistency between the models.
The benchmark test shows In-Context Earth performs best, with an MAE of 4.7 $^\circ$C.
The Stanford Thermal Model achieves an MAE of 6.4 $^\circ$C,  the model that leverages AlphaEarth embeddings achieves an MAE of 6.0 $^\circ$C, and Transparent Earth \cite{mazumder2025transparent} achieves an MAE of 5.8 $^\circ$C. 
Interestingly, universal kriging achieves an MAE of 5.1 $^\circ$C and outperforms these other advanced approaches. 
Only In-Context Earth outperforms universal kriging, corresponding to a 27\% reduction in MAE relative to the Stanford Thermal Model.

The benefit of in-context conditioning is not limited to the $M=20$ setting used for the main comparisons. 
A sensitivity analysis in the Supplementary Information shows that In-Context Earth remains highly accurate in the US even with only five in-context observations, achieving an MAE of 4.96 $^\circ$C compared with 4.75 $^\circ$C for the $M=20$ setting (Table \ref{tab:context_scaling}). 
This five-observation model still outperforms all US baselines in Table \ref{tab:mae_us}, including universal kriging, which is using 20 neighboring observations to inform its prediction.

\begin{table}
\centering
\caption{In-distribution quantitative performance comparison for the US shown as the MAE for the Stanford Thermal Model \cite{aljubran2024thermal}, AlphaEarth embeddings \cite{brown2025alphaearth,nakata2026subsurface}, Transparent Earth \cite{mazumder2025transparent}, and universal kriging.}
\label{tab:mae_us}
\begin{tabular}{lc}
\toprule
Model & MAE $^\circ$C \\
\midrule
Stanford Thermal Model & 6.4 \\
AlphaEarth embeddings & 6.0 \\
Transparent Earth & 5.8 \\
Universal Kriging & 5.1 \\
In-Context Earth & \textbf{4.7} \\
\bottomrule
\end{tabular}
\end{table}

\subsection*{Continental-scale shallow crust temperature throughout the US}
\begin{figure}
     \centering
     \includegraphics[width=\textwidth]{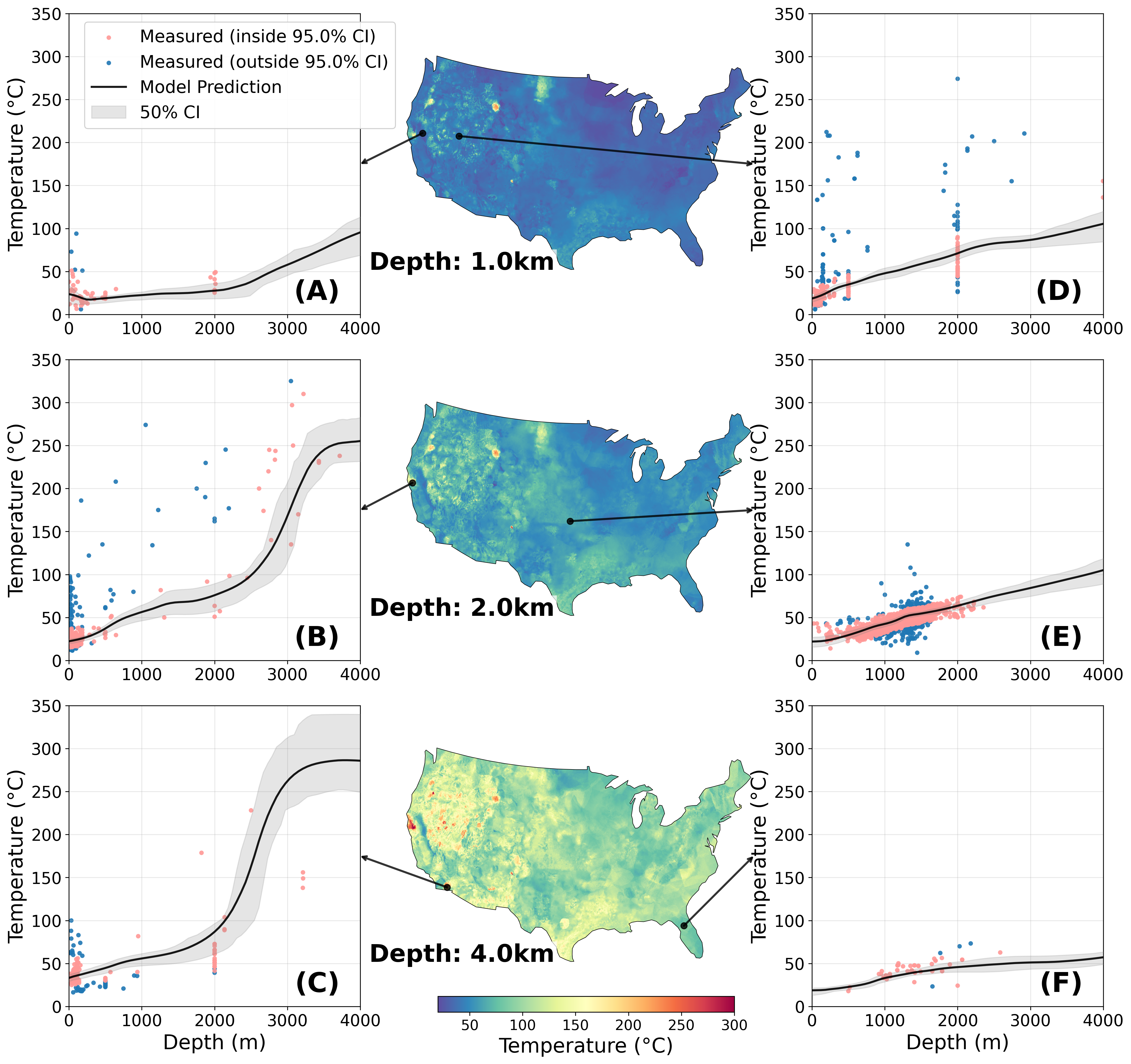}
     \caption{In the center are continuous temperature maps of the continental US resolved at 1 km, 2 km, and 4 km depths. Side panels show six locations, (A) northern Sierra Nevada, California, (B) Geysers geothermal field, California, (C) El Centro, California, (D) central Basin and Range, Nevada, (E) Ogallala aquifer, Kansas, and (F) karst aquifer, central Florida, each in a different tectonic and basin setting to highlight the depth resolution with borehole measurements within 100 km. The model prediction is shown as a function of depth with a 50\% confidence interval. In-situ measurements are shown inside (red dots) and outside (blue dots) the 95\% confidence interval.}
     \label{fig:usa_temperature}
 \end{figure}

In-Context Earth generates continuous temperature fields at 1, 2, and 4 km and shows a high-resolution, spatially complete view of the subsurface thermal regime (Figure \ref{fig:usa_temperature}).
The temperature mapping reveals spatial patterns that are broadly consistent with known tectonic and geothermal patterns, such as elevated temperatures across the western US and the relatively cool, stable thermal character of the central and eastern US.
Interestingly, the model reveals notable deviations from the first-order geologic-thermal relationship and expectations.
For instance, while the western US is typically characterized by smoothly varying high heat flow from tectonic extension \cite{parsons2006basin}, our model resolves internal heterogeneity and identifies localized thermal suppression and enhancement, which suggests the influence of regional high heat flow and the mitigating effects of fluid circulation in the upper crust.
At $\sim$1 km depth, the thermal patterns are relatively smooth and strongly influenced by surface boundary conditions.
Moving deeper to $\sim$4 km depth, the resolved thermal mapping becomes increasingly complex with the influence of deeper processes, including radiogenic heat production, variations in crustal thickness, and deep hydrothermal processes influencing the predicted borehole temperature.
This increasing complexity with depth suggests that the transformer-based conditioning mechanism is effectively capturing the nonlinear relationship between near-surface factors and deeper crustal heat transport, thus allowing the context-aware inference to resolve sharper thermal gradients.

To evaluate the In-Context Earth model predictions, we show site-specific depth profiles for six different geological settings in areas where sharp thermal gradients are present (see side panels in Figure \ref{fig:usa_temperature}).
Observational data points within the 95\% confidence interval are shown in red and there is good calibration of the uncertainty, which is quantitatively assessed later in the manuscript.
Outside this confidence interval the data are shown in blue.
Note that the observations are in the vicinity of the location where the temperature-depth profile is plotted, but are not exactly co-located, so discrepancies between the predicted curve and the data can reflect geothermal gradients in the region that the model captures.
These depth profiles highlight significant variability in measured temperatures within a radius of 100 km to provide a representative distribution of data points for a location.
In the northern Sierra Nevada the model correctly tracks the geothermal gradient and predicts temperatures of about 50$^\circ$C around 2 km depth.
Of particular interest are two geothermal regions in California, the Geysers geothermal field and El Centro. 
These areas contain a sharp thermal gradient that is captured by the model to depths greater than 3 km. 
At the Geysers, the model correctly predicts the thermal gradient increase to about 250$^\circ$C around 3 km depth but does underpredict some shallower, more localized heat anomalies. 
At El Centro, the model closely tracks the temperature-at-depth measurements and correctly predicts temperatures of about 225$^\circ$C at around 2.5 km depth.
In the central Basin and Range, the model predictions are similar to the Geysers geothermal field and only capture the lower threshold of the thermal profile, but correctly predict most of the measurements within the 95\% confidence interval.
In the central US at the Ogallala aquifer and the karst aquifer in central Florida the model performs very well and tracks the geothermal gradient at greater than 2 km depth. 
Despite the heterogeneity throughout the US, In-Context Earth effectively captures the thermal regime and the associated confidence intervals accurately capture model error represented by the local thermal variability.
Traditional mapping methods often obscure the sharp temperature gradients that are critical for resource identification.
In-Context Earth accounts for this by incorporating the high-frequency spatial heterogeneity inherent in geothermal temperature data and can thus resolve these features via the combined temperature prediction and uncertainty quantification.

\begin{figure}
    \centering
    \includegraphics[width=0.92\textwidth]{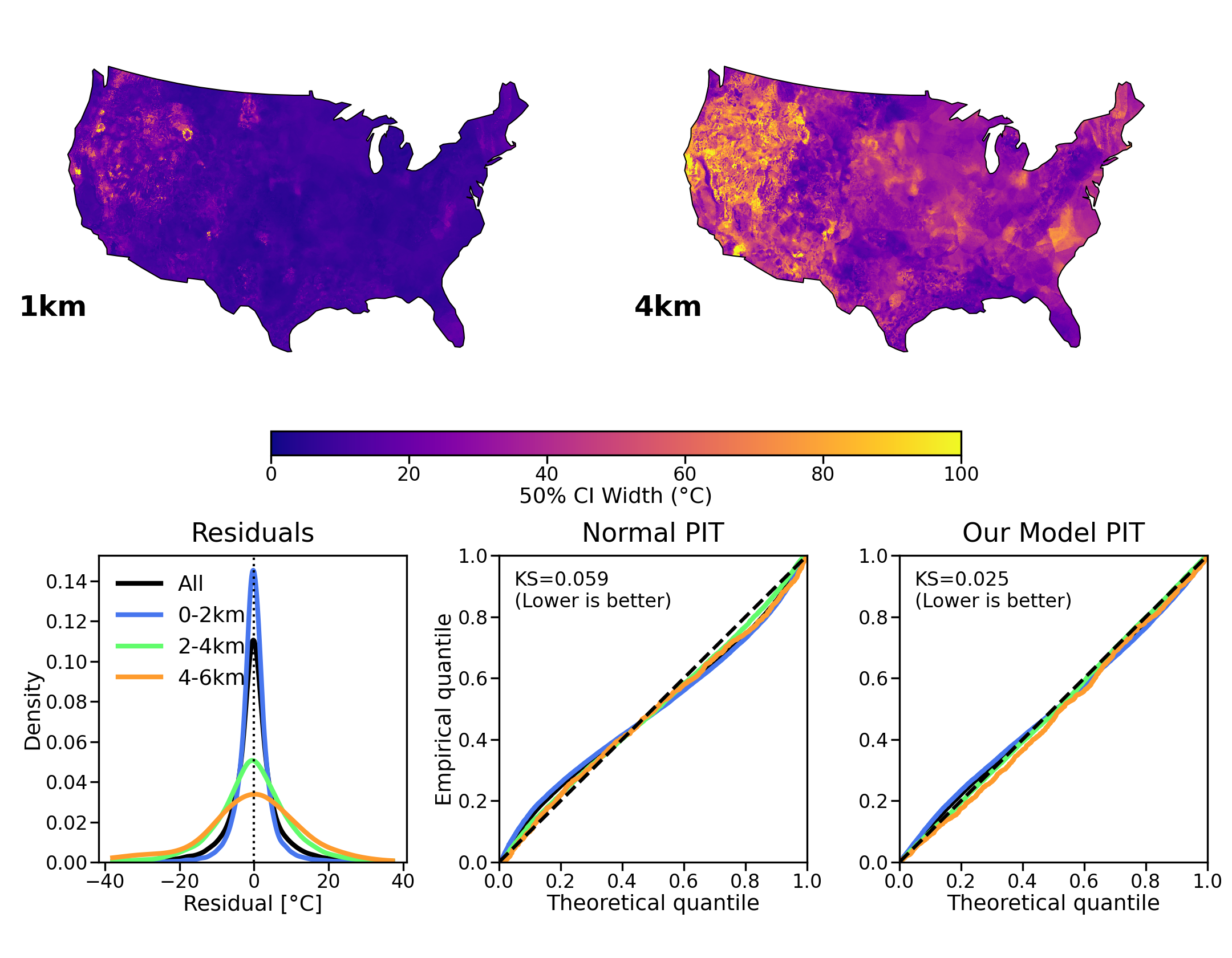}
    \caption{(Top) Prediction uncertainty (50\% confidence interval) at 1 and 4 km depth. (Bottom-Left) Residual error distributions are shown as a function of depth, reflecting an accurate model that becomes increasingly uncertain as depth increases. (Bottom-Right) Q-Q plot quantifies the difference between the observed residual distribution and the residual distribution predicted by the model. The curves follow the dashed line indicating similar distributions. (Bottom-Middle) A normal approximation of the residual is used given the approximately Gaussian distributions. Included in the Q-Q plot is the Kolmogorov-Smirnov statistic, with In-Context Earth improved by more than a factor of 2 when compared to the normal approximation.}
    \label{fig:usa_uncertainty}
\end{figure}

The spatial distribution of predictive uncertainty is quantified using the width of the 50\% confidence interval and provides information on the reliability of the inferred thermal fields (Figure \ref{fig:usa_uncertainty}).
Lower uncertainties are associated with a high density of borehole observations (Figure \ref{fig:data_us_temperature_observations}) and a systematic increase in the uncertainty is observed with depth.
In densely sampled regions the confidence intervals are relatively narrow, reflecting lower uncertainty at shallow depths and where data are abundant.
At 4 km depth the map shows much more variability and the largest uncertainty in the tectonically active western US. 
The MAE increases from 3.1 $^\circ$C in the 0-2 km range, to 7.9 $^\circ$C in the 2-4 km range, and to 11.2 $^\circ$C in the 4-6 km range.
The increased uncertainty also appears to correlate with structural complexity in the western US.
In these areas, sharp lateral thermal gradients and complex tectonic histories make inference challenging, particularly where thermal anomalies may exist between sparse observation points.
This uncertainty structure underscores the importance of a probabilistic framework, ensuring that high-resolution geothermal maps are interpreted alongside their corresponding confidence levels.

Residual distributions, i.e., the distribution of true temperature minus the predicted temperature, for the US most clearly show the increasing variance of the residuals as a function of depth (bottom left of Figure \ref{fig:usa_uncertainty}).
As the model predictions move deeper from 0-2, 2-4, and 4-6 km, the standard deviation systematically increases.
All are centered near zero, reflecting little or no bias in the model’s predictions.
This depth-dependent trend reflects the diminishing influence of near-surface constraints, the increasing complexity of deeper crustal heat transport processes, and increasing data paucity at depth, especially since only 3\% of the measurements are deeper than 4 km.
However, even at 6 km where observations are becoming very sparse, the error distributions remain approximately symmetric and Gaussian.
A similar analysis focusing on distance to the in-context temperature observations shows broadly similar patterns (Figure \ref{fig:error_vs_distance}).
The lowest residuals are observed when the distance to neighbors is less than 5 km.
Residuals gradually increase with the neighbor distance, and uncertainty remains well calibrated when the in-context data are either near or far.

The probability integral transform (PIT) and Q-Q analyses on the US validation set (bottom center and right of Figure \ref{fig:usa_uncertainty}) show the statistical validity and calibration of the inferred thermal fields.
The PIT-based Q-Q plot shows close agreement between the empirical PIT quantiles and the theoretical uniform quantiles, indicating that the model’s predictive distributions are well calibrated.
Like the accuracy of temperature predictions, uncertainty estimation becomes more challenging at depth where data are sparser, but remains broadly calibrated.
Quantitatively, the Kolmogorov-Smirnov (KS) statistic is 2.5\%, less than half the 5.9\% achieved by a standard normal assumption baseline, indicating the model is performing well.
The low KS value confirms that the probabilistic output accurately reflects the error structure.
These results suggest that the transformer-based conditioning mechanism avoids the over- or under-confidence problem that is common with many spatial interpolation techniques that include uncertainty quantification.
This uncertainty calibration ensures that the estimates provided by In-Context Earth are reliable for risk assessment and decision-making in geothermal exploration.

\subsection*{Few-shot generalization to unseen regions}
\begin{figure}
    \centering
    \includegraphics[width=\textwidth]{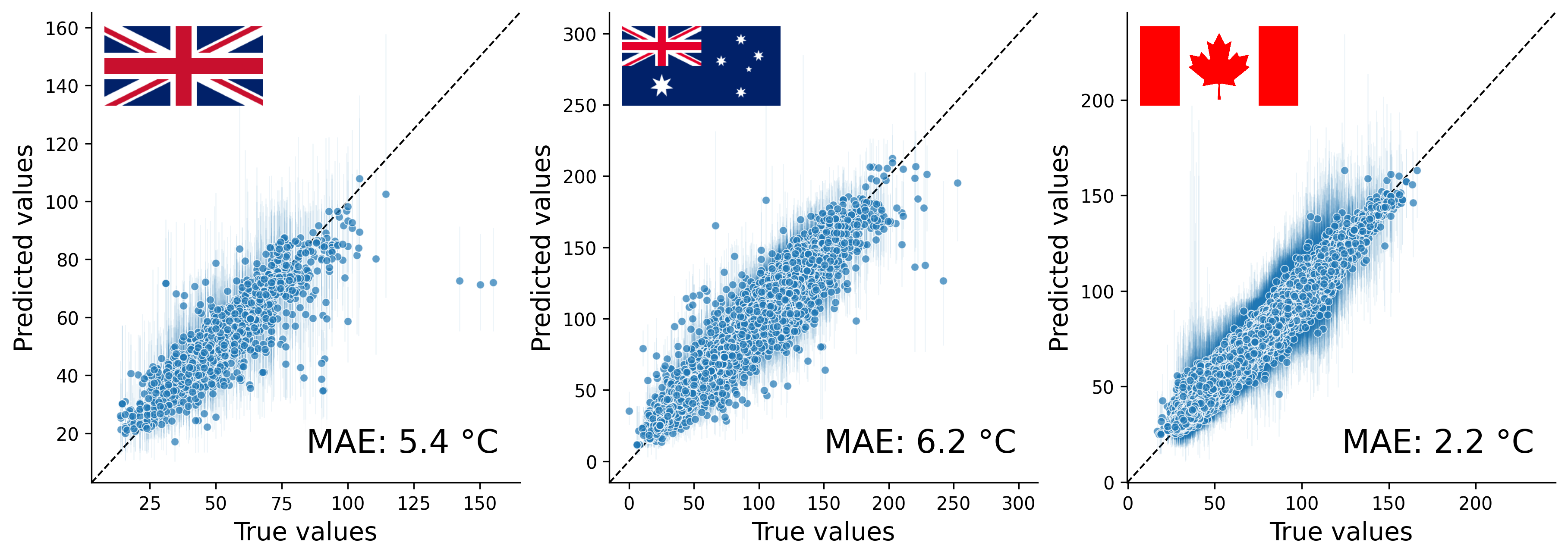}
    \caption{Parity plots showing the In-Context Earth predicted values and the observed temperatures for the three out-of-distribution regions. The light blue lines show the model's 95\% confidence interval for each prediction.}
    \label{fig:ood_performance}
\end{figure}

In-Context Earth is applied to three regions outside the US to evaluate generalization to new regions. 
This reflects the ability to utilize transferable representations of subsurface heat transport rather than regional memorization and test the hypothesis that the relationship between sparse observations and the broader crustal thermal regime can be learned by the model.
We evaluate the out-of-distribution performance on a region in western North America in Alberta, Canada and on other continents using two datasets from Australia and the United Kingdom (UK) (see Figure \ref{fig:ood_performance}).
These regions represent a spectrum of tectonic histories, stratigraphic sequences, and sampling densities.
The model was applied to these data with no retraining, finetuning, or region-specific parameter adjustments, but it did use 20 local temperature observations in context when being evaluated in these regions (see Methods).
Despite the significant geological differences between the training data from the US and these unseen regions, the model accurately resolves regional thermal structures, suggesting that the transformer-based conditioning mechanism effectively generalizes beyond the geographic extent of its initial training data.

We also compare the out-of-distribution performance of In-Context Earth, Transparent Earth, and universal kriging by examining their performance on data from Alberta (Canada), Australia, and the UK (Table \ref{tab:mae_ood}).
Compared to universal kriging, In-Context Earth reduces errors by between 39\% and 78\%.
Notably, the worst out-of-distribution performance for In-Context Earth is 6.2 $^\circ$C in Australia, which still slightly outperforms the Stanford Thermal Model’s in-distribution score.
Australia is a stringent transfer test because its thermal field includes strong intraplate heterogeneity and basin-scale thermal complexity, especially in the Cooper-Eromanga system (Supplementary Information Section \ref{sec:aus-thermal-complexity}).
Note that the Stanford Thermal Model cannot be applied in these other regions and relies on a wide variety of data that is not readily available.
Retraining the Stanford Thermal Model for a new region requires extensive data collection, since the Stanford Thermal Model relies on many data sources (seismic velocities, crustal thickness, geochemistry, etc.).
The model based on AlphaEarth embeddings could potentially be used to perform inference in these regions, but scores were not reported by those authors \cite{nakata2026subsurface}.
The out-of-distribution performance illustrates that the In-Context Earth model is not memorizing regional patterns but is learning to interpret sparse contextual observations in a way that generalizes.
The most extreme UK and Australia residuals are associated with data-quality concerns, including measurements while drilling, assigned temperatures, inconsistent nearby measurements, and depth errors.
In contrast, Canada does not contain similarly extreme residual outliers, and its largest residual is only slightly outside the model's 95\% confidence interval.
These observations were retained in the reported test metrics to avoid selectively filtering difficult cases; details are provided in Table \ref{tab:bht_largest_outliers}.
The US training data provides the foundational representations of heat transport and the model effectively adapts these patterns to the thermal regimes of Australia, Canada, and the UK.

\begin{table}
\centering
\caption{Out-of-distribution quantitative performance comparison across unseen continental regions shown as the MAE for universal kriging, Transparent Earth, and In-Context Earth in Alberta, Australia, and the UK. The Stanford Thermal Model and AlphaEarth-embedding baseline are not included because scores for these methods are not reported in these regions.}
\label{tab:mae_ood}
\begin{tabular}{lccc}
\toprule
Model &
\shortstack{Alberta} &
\shortstack{Australia} &
\shortstack{UK} \\
\midrule
Universal Kriging & 9.2 & 28.5 & 8.9 \\
Transparent Earth & 15 & 30 & 9 \\
In-Context Earth (Ours) & \textbf{2.2} & \textbf{6.2} & \textbf{5.4} \\
\bottomrule
\end{tabular}
\end{table}

\begin{figure}
    \centering
    \includegraphics[width=\textwidth]{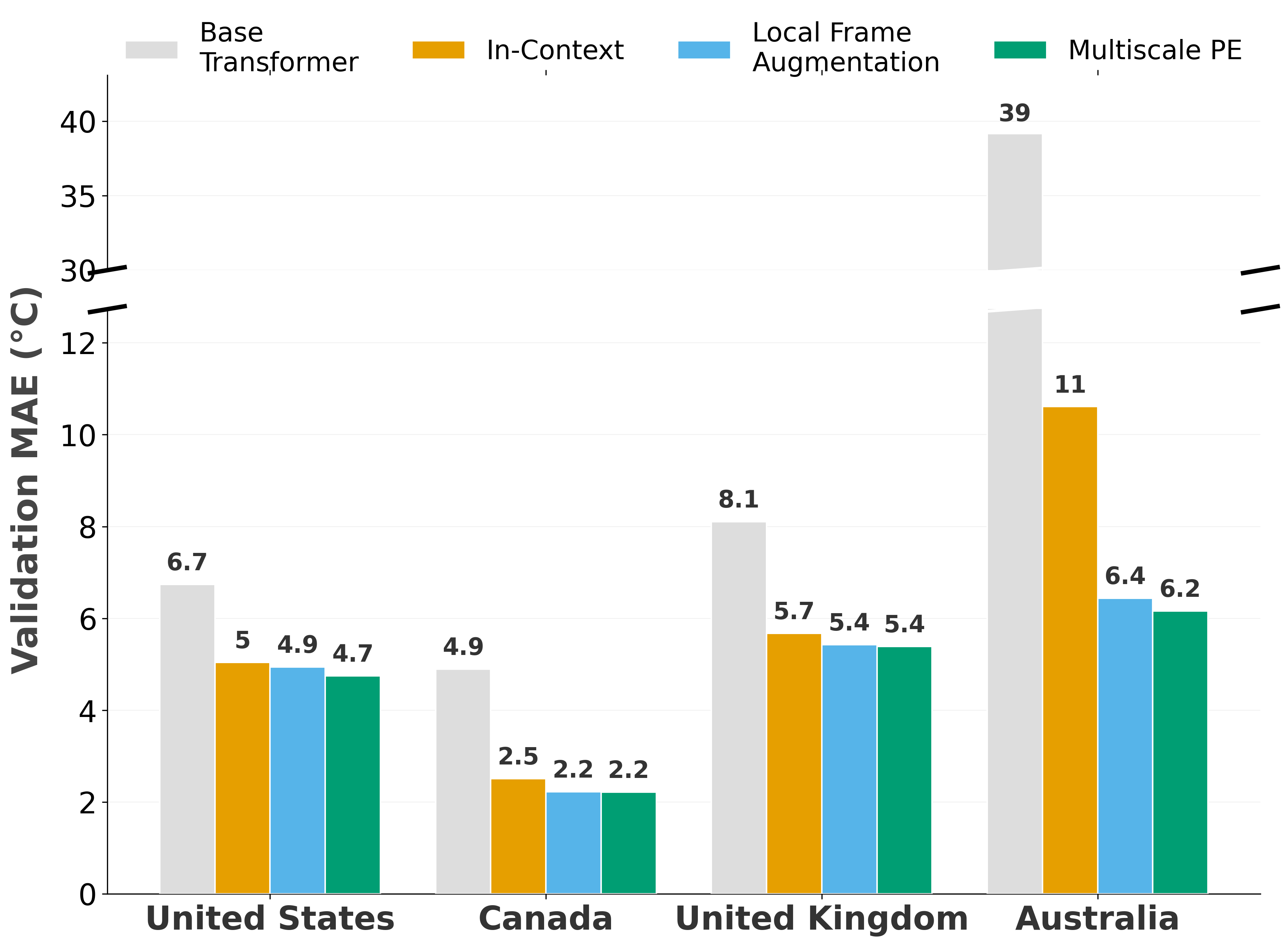}
    \caption{Ablation test results for four regions.
    The improvement in MAE as these key features are added demonstrates the ability of our model to generalize to out-of-distribution regions (Canada, the UK, and Australia).
    These features include: base transformer, in-context learning, local frame augmentation that is a form of Earth-tailored data augmentation, and multiscale positional encodings.}
    \label{fig:ablations}
\end{figure}

The ability of In-Context Earth to maintain high predictive accuracy in these unseen domains suggests that the underlying physics of crustal heat transport is learnable and transferable.
While the US training set encompasses a range of thermal regimes, the successful transfer to regions with fundamentally different stratigraphic and tectonic characteristics indicates that the model is learning how sparse temperature observations relate to subsurface thermal processes.
The results imply that the model learns a prior for heat transport that can be dynamically updated by local observations.
This suggests that the relationship between sparse temperature data and geothermal structure is globally consistent.
To evaluate the contribution of key model features that facilitate strong out-of-distribution performance, we performed a systematic ablation study on the model architecture (Figure \ref{fig:ablations}).
By adding the transformer-based conditioning mechanism, the multiscale spatial encoding, and our Earth-tailored data augmentation one at a time, we quantified the impact of each on both in-distribution performance and, more crucially, out-of-distribution generalization.
We focused on the predictive accuracy as these features were sequentially included.
This process identified the specific model design required for In-Context Earth to effectively transition from regional memorization to a transferable understanding of crustal heat transport.
The capacity for global generalization is the ability to treat sparse local measurements as a location-agnostic conditioning set.
This shifts the modeling approach to a conditioning paradigm, where the model learns the underlying geological contexts and deploys transferable representations of heat transport.
We identify in-context learning, Earth-tailored data augmentation to ensure coordinate-invariant learning, and multiscale positional encodings to resolve clustered local information as key to strong model performance.
Removing these specific architectural choices shows the performance on continents not seen during training can degrade by nearly an order of magnitude (e.g., in Australia), supporting that learned conditioning on geological context is the primary driver of robust global generalization.

\subsection*{Internal representations encode geologically meaningful properties}

\begin{figure}
    \centering
    \includegraphics[width=\textwidth]{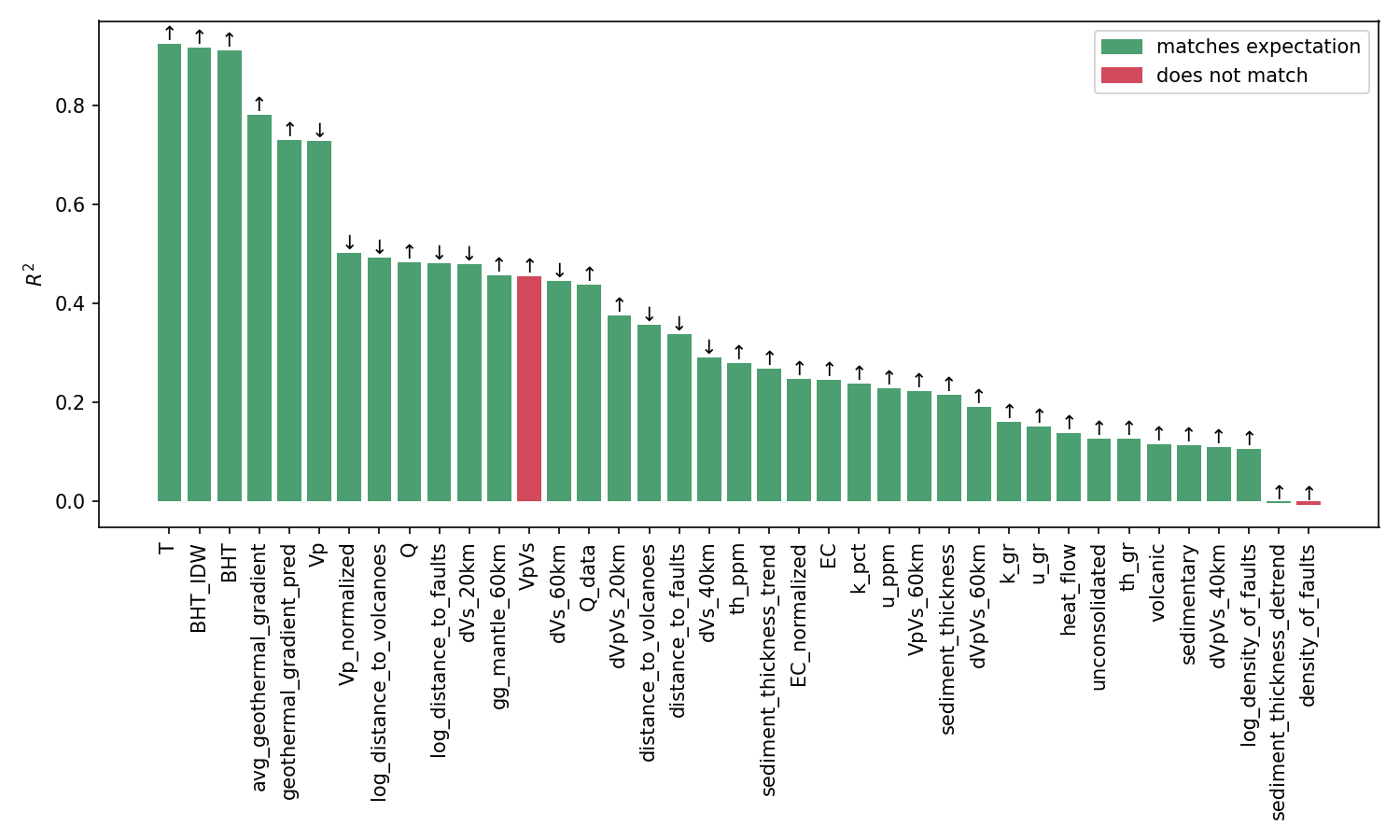}
    \caption{The strength of the representation ($R^2$) for 39 subsurface features where there is a physical expectation about the relationship (either direct or inverse) between the variable in the dataset and geothermal temperature. The arrows above the bar indicate whether the relationship is direct ($\uparrow$) or inverse ($\downarrow$). The color indicates whether the model's use of this representation agrees with the expectation -- green for agreement and red for disagreement. The model's use of the representations agrees with the physical expectation in 37 of 39 cases with one of the disagreements coming from a variable where the $R^2$ value is nearly zero (slightly negative), indicating the model does not have a meaningful representation of this variable.}
    \label{fig:representation_usage}
\end{figure}

The out-of-distribution performance suggests that the model leverages internal representations that capture broader geological context rather than acting only as a sophisticated spatial smoother.
We test this hypothesis by analyzing whether the final-layer query-token activations encode information about a wide variety of subsurface properties including geophysical (e.g., seismic velocities) and geochemical (e.g., elemental concentrations).
This approach is inspired by the representation interventions performed on the OthelloGPT model \cite{li2022emergent}.
The data used are the Stanford Thermal Model training dataset.
The In-Context Earth model does not see these data during training -- it sees only temperature data.
The reported metric is the $R^2$ value of a linear probe trained to predict Stanford Thermal Model variables from the final-layer query representation (Figure \ref{fig:representation_usage}).
We focus on variables for which there is a clear physical expectation about whether increasing the variable should (all else being equal) increase or decrease geothermal temperature, and we indicate this expectation with an arrow ($\uparrow$ or $\downarrow$).
There are 39 such variables (Figure \ref{fig:representation_comparison} contains information on all variables, including ones where there is no clear relationship to temperature).
The $R^2$ value quantifies how strongly the variable is linearly represented and it provides a set of targets for testing whether the model uses these representations in a physically meaningful way.
We identify a direction in representation space associated with increasing that variable via the linear probe and apply controlled shifts in that direction to the final-layer query activation during a forward pass.
By shifting in that direction, we can essentially trick the model into thinking that, e.g., the crust is thicker (to name one of the 39 properties) than it really is at the query location.
We then measure the induced change in the expected temperature prediction, $\Delta T$.
Strikingly, the induced temperature response agrees with physical expectations in 37 out of 39 cases.
One of the two mismatches corresponds to a variable whose linear probe has essentially no signal ($R^2\approx 0$ and slightly negative), indicating that the model does not meaningfully represent that quantity.

Agreement rates must be interpreted relative to baselines.
If a model had no usable representation of these subsurface properties, then the sign of an intervention would match the expected direction for only about half of the variables, 19 or 20 out of 39.
The base transformer ablation without in-context learning, Earth-tailored data augmentation, or multiscale positional encodings achieves agreement on only 26 variables.
The base transformer model learns representations that often have higher $R^2$ values (see Figure \ref{fig:representation_comparison}), but fails the intervention test of whether these representations are used in physically sensible ways.
This contrasts with the performance of In-Context Earth which uses the representations in ways that align with physical intuition, providing evidence that In-Context Earth learns a crude but geologically meaningful internal world model in the sense of learned latent representations of the environment \cite{ha2018recurrent}.
The final-layer query representation linearly encodes multiple subsurface properties, and manipulating these inferred factors yields temperature responses that align with known physical relationships.
A nuisance-controlled analysis (Figure \ref{fig:nuisance_probe}) shows that the final-layer query activation adds predictive information beyond query location, local temperature summaries, neighbor-distance statistics, depth relationships, and the model's own temperature prediction, improving held-out $R^2$ for 73 of 80 Stanford Thermal Model variables.
This indicates that the probe results are not due to, e.g., correlations between temperature and other subsurface properties, but reflect additional geological information encoded in the model's internal representation.
In cases where the relationship between temperature and a subsurface property is not known, this interpretability approach could potentially be used to generate a plausible hypothesis for the relationship.

\section*{Discussion}

The In-Context Earth model presents a new paradigm in how the shallow crust thermal regime and other subsurface quantities of interest can be inferred using advanced deep learning techniques. 
This approach shifts from purely localized spatial interpolation to an in-context learning framework that is analogous to the textual prompts provided to large language models.
Treating sparse borehole observations as a ``geological prompt'' in combination with the model's training on many small-data examples allows In-Context Earth to maximize the value of minimal local information.
This in-context learning capability allows the model to leverage a global prior of crustal heat transport, learned directly from the sparse measurements throughout the US, and apply those representations to other regions.

In-Context Earth has implications for geothermal exploration in regions where data paucity has previously rendered high-fidelity mapping impossible.
The model learns a complex, nonlinear relationship between observations and the surrounding geology instead of relying on regularized spatial assumptions.
This is evidenced by the ability to resolve relatively sharp thermal gradients in complex provinces like the western US.
Our modeling framework more effectively captures the subtle signatures of advective heat transport and heterogeneous crustal properties that can be difficult to resolve with idealized physical equations alone.
Notably, the Stanford Thermal Model \cite{aljubran2024thermal} uses physics-informed machine learning and applies a constraint based on Fourier's law of heat conduction that treats heat transport as a purely diffusive process and neglects advective heat transport that can produce sharp gradients.
Rather than using physics-informed machine learning, In-Context Earth is aligned with the Bitter Lesson \cite{sutton2019bitterlesson}, which argues that scaling computation with search and learning outperforms approaches that encode our assumptions directly into a model.
At inference time, search selects the most relevant observations to provide as context, and following Tobler's law (``near things are more related than distant things'') \cite{tobler1970computer}, we retrieve nearby data for each query location.
Learning then trains the model to interpret this local evidence and predict the temperature distribution implied by the surrounding geological context.
In this sense, In-Context Earth operationalizes the Bitter Lesson for subsurface prediction by scaling search and learning to discover the geological structure that sparse observations imply.

In-Context Earth also moves geothermal mapping from static assessment toward operational decision support.
During directional drilling, the model could use in-situ temperature measurements as they are collected and predict the temperature of the surrounding volume in real time.
The model is also lightweight enough for field deployment.
Because it has fewer than one million parameters, it can run in the field without the high-powered GPUs typically associated with modern deep learning.
As new downhole measurements are collected, In-Context Earth can update the local temperature field and help guide drilling in hotter, more promising directions.

Despite these advances, In-Context Earth is subject to limitations.
Predictive uncertainty systematically increases with depth, reflecting both the increasing sparsity of training data and the increasing contribution of local effects produced by deep subsurface processes to the thermal profile.
While the model demonstrates remarkable transferability to diverse regions, the overall performance is bounded by the representativeness of the temperature patterns it has encountered during training.
The strong performance in three out-of-distribution regions does provide confidence that the model can generalize, but it remains possible that some out-of-distribution regions could confound the model.
The requirement for in-context data also means the model is not a tool for prediction in a vacuum.
Rather, it is a mechanism for using local evidence to inform geological understanding by exploiting the prior learned from the training data, similar to Bayesian methods \cite{agarwal2025bayesian}.

This context-aware approach is broadly applicable to many Earth science problems where observations are sparse and irregularly sampled.
The current study focuses on providing temperature values in-context. 
Enhancing the in-context capabilities of these models by expanding the types of geological data provided during inference is still needed.
The highly functional in-context learning approach developed here is adaptable and could ingest diverse data types in varying quantities to better characterize the subsurface for a suite of geophysical measurements.
Examples include location-specific information on crustal thickness, rock type, and local stratigraphy alongside temperature measurements that may allow the model to better resolve complex thermal regimes and more finely resolve temperature gradients.
Similarly, providing high-resolution local data, such as information on stress, permeability, or geochemical concentrations, to a model pretrained with global datasets can be used to obtain high-fidelity inference across a wide array of geological domains.
A new model design capable of integrating such multimodal inputs would represent a significant step toward a more holistic characterization of the shallow crust.
Expanding the scope of in-context learning in this way provides a scalable path for subsurface characterization across the Earth sciences.

\section*{Methods}

\subsection*{Borehole data sources and quality control}
The primary dataset utilized for training our model consists of a comprehensive compilation of borehole temperature measurements from the contiguous US \cite{aljubran2024thermal,aljubran2024stanford}.
To evaluate the global transferability of In-Context Earth, we curated independent datasets from three geographically and geologically distinct regions: the UK, Australia, and Alberta, Canada.
The UK data were obtained from the UK Onshore Geophysical Library (UKOGL) archives (\url{https://ukogl.org.uk/archives/}), which provide a record of subsurface thermal gradients in the UK.
Australian temperature data were retrieved from the OZTemp Well Temperature dataset on the Geoscience Australia portal \cite{holgate2010oztemp}.
The Canadian dataset was sourced from the Alberta Geological Survey \cite{brinsky2022subsurface}.
We excluded records with missing or non-physical depth and temperature values.
The spatial and depth distributions of the temperature observations used for the US, Alberta, Australia, and the UK are shown in Figures \ref{fig:data_us_temperature_observations}-\ref{fig:data_uk_temperature_observations}.
Furthermore, we integrated digital elevation models from the ETOPO1 Global Relief Model \cite{amante2009etopo1} to append surface elevation as a localized feature, providing critical context for shallow thermal boundary conditions.

\subsection*{Model design}
The In-Context Earth model is a modified encoder-only Transformer that is designed to ingest a variable number of context tokens representing borehole measurements sampled from the surrounding geological region to inform the prediction at a specific target location.
The architecture utilizes a series of Transformer encoder layers with multi-head self-attention.
This mechanism allows each context token to attend to all others, building a representation of the local thermal environment.
The target prediction location is introduced as a special query token, where the temperature component is masked.
The model processes the entire sequence, and the final hidden state of the query token is passed to a linear layer to predict the probabilities of being in each temperature bin.
Each logit defines the probability that the temperature exceeds a threshold associated with the temperature bins.
The output layer ensures that the predicted probabilities represent a valid, monotonically consistent thermal distribution, i.e., the probability of exceeding a high threshold is less than the probability of exceeding a lower threshold.

To represent the spatial relationships in the temperature distributions, we employ a multiscale spatial encoding mechanism.
Depth values were standardized to zero mean and unit variance based on the global distribution of the training set.
Spatial coordinates are encoded as trigonometric representations ($\sin(\omega_i\mathrm{lat}), \cos(\omega_i\mathrm{lat}), \sin(\omega_i\mathrm{lon}), \cos(\omega_i\mathrm{lon})$).
Specifically, we apply a range of multiscale frequencies $\omega_i = 2^i \pi$ for $i \in \{0, \dots, S-1\}$, which allows the model to resolve both broad continental trends and tightly clustered local information while keeping the floating point representation appropriately scaled.
The $\sin$ and $\cos$ values are concatenated with depth and surface elevation to provide a positional encoding.
Target temperature values were similarly normalized using the mean and standard deviation of the training distribution.
We treat temperature predictions as a classification task, which enabled robust uncertainty quantification using a conditioning mechanism where target temperatures are binned into $K$ discrete intervals.
Temperatures are put into bins representing temperature ranges with a one-hot encoding.
We use 128 temperature bins that span 8 standard deviations in the temperature space.
The model is tasked with predicting the probability distribution across these bins.
This allows the model to represent non-Gaussian and other non-parametric uncertainty structures, which are common in geologically complex regions.
Viewed through a language-model lens, the $K$ bins act as a small vocabulary and the transformer is trained to predict the masked temperature token conditioned on the context sequence.
The input to the transformer consists of a sequence of tokens, where each token is a concatenated vector of the spatially encoded features ($X_i$) and its corresponding binned temperature ($y_i$).

To facilitate global generalization and ensure coordinate invariance, we incorporate Earth-tailored data augmentation directly into the forward pass.
During training, the spatial coordinates of each context set are subjected to instance centering (so the centroid of the in-context points and the query point is at the origin in latitude and longitude space) and random rotation around the vertical (depth) axis.
This prevents the model from short-circuiting the learning process by memorizing a function that maps latitude, longitude, and depth to a specific temperature.
This forces the model to learn representations based on the relative spatial configuration and thermal gradients of the boreholes rather than memorizing temperatures at absolute geographic coordinates.

\subsection*{Training regime and validation approach}
The training regime is structured to facilitate a transition from absolute spatial memorization to a conditional inference paradigm.
The model is trained on borehole temperature measurements from the contiguous US, representing a diverse range of tectonic and geothermal provinces.
To implement the in-context learning framework, we employ a stochastic training strategy where each sample is constructed by dynamically selecting a target point and a corresponding context set of $M$ neighboring boreholes.
We use a random 90/10 training/validation split for the US data in the main text.
This allows us to use held-out continental regions as test datasets, which is a strong form of spatial blocking.
During both training and validation, the training data are used to provide in-context data to the model.
Because in-context learning conditions predictions on nearby observations, a possible concern is that the context set could include measurements from the same well as the target point, which would blur the distinction between valid usage of in-context information and data leakage.
Results from additional splits to address this issue more specifically are described in the Supplementary Information and Table \ref{tab:mae_other_splits} and do not substantially change the model's performance.
This issue is of little concern here because more than 95\% of wells contain only a single temperature measurement in our US dataset, so it is rarely possible to provide context from the target well.
Of course, the target measurement itself is never included in the context set.

Central to In-Context Earth is an in-context conditioning approach.
During both training and inference, each target prediction is conditioned on a set of $M$ context points.
These context points are dynamically selected using a KD-Tree search based on the distance to identify the nearest neighboring data points.
By treating these neighbors as geological tokens, the transformer-based architecture learns to interpret the target point's thermal state relative to its surrounding environment.
To enhance the robustness to varying data densities, we implement a stochastic downsampling strategy.
We initially query a pool of $2M$ neighbors and randomly select $M$ points.
A value of $M=20$ was used throughout the main paper to serve as the context, but Table \ref{tab:context_scaling} contains information on performance for other values of $M$.

During each training iteration, the model is presented with a sequence of $M+1$ tokens.
The first $M$ tokens consist of context points $(X_i, y_i)$, where $X_i$ contains spatially encoded features (latitude, longitude, depth, and elevation) and $y_i$ is the quantized temperature.
The final token is the target query $(X_{target}, 0)$, where the quantized temperature portion of the token is set to all zeros and a learnable array is added to this final token.
The model is optimized to predict the discrete probability distribution of the target temperature using an ordinal cross-entropy loss.
This loss function is designed for the quantized output, as it penalizes predictions more heavily as they deviate further from the correct temperature bin.

\subsection*{Baseline models}
To establish a geostatistical baseline, we implement a universal kriging workflow that utilizes the 20 nearest neighbors (comparable to the $M=20$ neighbors used by In-Context Earth).
This baseline treats the subsurface thermal regime as a spatially correlated process, where predictions are derived from a best linear unbiased estimator of the local neighborhood.
To account for the dominant vertical temperature gradient, the baseline incorporates a linear depth-only trend ($T = a + bz$) fitted to the training data.
We also evaluated a geostatistical model with a full 3D trend, but it performed very poorly outside the US.
This vertical trend is removed from the observations, and kriging is performed on the resulting residuals.
The local neighborhood search is conducted in a 3D scaled coordinate space (latitude, longitude, and depth $\times$ 40), so that vertical and horizontal distances are weighted differently for geothermal correlation.
A scaling factor of 40 in the vertical direction significantly improved predictive performance compared to an isotropic model and the accuracy of the model was insensitive to the scaling factor when it was approximately 40 -- the MAE remained 5.1 when the scaling factor was in the range of 30 to 50.
For each test point, we estimate a global exponential variogram (using a nugget, partial sill, and range parameter) by fitting the model to an empirical semivariogram generated from a random subsample of 300,000 training point pairs.
The kriging system is then solved locally for each prediction point, and the previously removed depth trend is added back to produce the final predicted temperature.
This configuration ensures that the baseline benefits from similar local information as our model, allowing for a direct comparison between traditional geostatistical interpolation and our transformer-based conditioning mechanism.
A version of Transparent Earth \cite{mazumder2025transparent} was trained with the addition of the US temperature data used here and validated on the same data from the UK, Alberta, and Australia.
In the US, we also compare to a recently reported score for a model based on AlphaEarth embeddings \cite{nakata2026subsurface,brown2025alphaearth}.

\subsection*{In-distribution and out-of-distribution evaluation protocols}
The evaluation of In-Context Earth is conducted across two distinct protocols: in-distribution (ID) validation within the US and testing on unseen continental regions (Australia, Canada, and the UK).
In the ID protocol, we assess the model on a withheld test set (not used during training) using 10\% of the temperature data in the US.
During both training and ID validation, the in-context information is from the training dataset in the US.
Performance is quantified using MAE.

To evaluate the transferability of the learned geological representations, the model is applied to independent datasets from Australia, Alberta, and the UK with no retraining or region-specific parameter tuning.
This frozen-weight protocol tests the hypothesis that the model has learned a globally-valid prior of crustal heat transport that can be dynamically updated by local observations.
During inference in these transfer regions, the model is provided with $M=20$ local borehole measurements as a conditional context set excluding the location where the prediction will be made.

The statistical reliability of these predictions is verified using the PIT, which evaluates whether the observed values are calibrated under the predicted cumulative distribution.
For calibrated predictions, PIT values should approximately follow a uniform distribution.
The Kolmogorov-Smirnov statistic is used to quantitatively evaluate the match between these two distributions.

\subsection*{Representation probing and interventions}
To quantify what information is present in the model's internal activations, we run the trained transformer on the temperature-in-context inference task and extract the activation of the query token at the final transformer layer.
For each target variable in the Stanford Thermal Model dataset, we fit a linear ridge-regression probe from this activation to the target variable and evaluate probe performance using a held-out test split, reporting $R^2$.

To test whether the model uses an inferred variable in a way consistent with known physical relationships (rather than merely encoding the variable), we perform representation interventions inspired by OthelloGPT \cite{li2022emergent}.
Specifically, after fitting a probe for a given variable, we treat the probe weight vector as a direction in representation space and add controlled shifts to the final-layer query activation during the forward pass.
For each intervention strength, we re-run the model and compute the change in the expected temperature prediction, $\Delta T$.
The sign of $\Delta T$ is then compared to the expected physical direction for that variable (e.g., increasing uranium concentration should increase temperature or increasing crustal thickness should decrease temperature).

\section*{Acknowledgements}
This material is based upon work supported by the U.S. Department of Energy, Office of Science, Office of Basic Energy Sciences, Geosciences program under Award Numbers LANLECA1 (support for DO, BS, JK) and LANLE3CB (support for CWJ and PL). LANL's Laboratory Directed Research \& Development program's Artimis project also supported DO, JES, PL, SM, BS, ND, EL, and HV.

\sloppy
\bibliographystyle{unsrt}
\bibliography{refs}
\fussy

\appendix
\section{Supplementary Information}

\setcounter{figure}{0}
\renewcommand{\thefigure}{S\arabic{figure}}

\setcounter{table}{0}
\renewcommand{\thetable}{S\arabic{table}}

\subsection{Training Details}

\begin{table}[H]
\centering
\small
\resizebox{\textwidth}{!}{%
\begin{tabular}{lcccc}
\toprule
Hyperparameter & Base Transformer & In-Context & Local Frame Augmentation & Multiscale PE \\
\midrule
Random seed & 0 & 0 & 0 & 0 \\
Batch size & 1024 & 1024 & 1024 & 1024 \\
Epochs & 120 & 120 & 120 & 120 \\
Optimizer & AdamW & AdamW & AdamW & AdamW \\
Learning rate & $3\times10^{-4}$ & $3\times10^{-4}$ & $3\times10^{-4}$ & $3\times10^{-4}$ \\
Weight decay & $1\times10^{-6}$ & $1\times10^{-6}$ & $1\times10^{-6}$ & $1\times10^{-6}$ \\
Gradient clipping & $\|g\|_2 \leq 10$ & $\|g\|_2 \leq 10$ & $\|g\|_2 \leq 10$ & $\|g\|_2 \leq 10$ \\
Schedule & 100-step warmup; cosine decay & 100-step warmup; cosine decay & 100-step warmup; cosine decay & 100-step warmup; cosine decay \\
\midrule
Target discretization & 128 bins over $[-8,8]$ & 128 bins over $[-8,8]$ & 128 bins over $[-8,8]$ & 128 bins over $[-8,8]$ \\
Loss & Ordinal cross entropy & Ordinal cross entropy & Ordinal cross entropy & Ordinal cross entropy \\
\midrule
Architecture & Encoder-only transformer & Encoder-only transformer & Encoder-only transformer & Encoder-only transformer \\
Hidden dimension & 128 & 128 & 128 & 128 \\
Attention heads & 4 & 4 & 4 & 4 \\
Transformer layers & 4 & 4 & 4 & 4 \\
Feed-forward dimension & 512 & 512 & 512 & 512 \\
\midrule
Main context points & \textbf{0} & \textbf{20 from pool of 40} & \textbf{20 from pool of 40} & \textbf{20 from pool of 40} \\
Normalization & Enabled & Enabled & Enabled & Enabled \\
\midrule
Local Frame Augmentation & \textbf{Disabled} & \textbf{Disabled} & \textbf{Enabled} & \textbf{Enabled} \\
Positional encoding scales & \textbf{1} & \textbf{1} & \textbf{1} & \textbf{16} \\
\bottomrule
\end{tabular}%
}
\caption{Training details for the step-by-step ablation study. The same transformer backbone,
optimization procedure, target discretization, and normalization are used throughout. All models have $\sim$830k parameters. Bold entries
indicate the components changed by the ablation sequence: adding local in-context observations,
then local-frame augmentation, and finally multiscale positional encodings.}
\label{tab:ablation-training-details}
\end{table}

\subsection{Performance versus amount of in-context data}

\begin{table}[H]
\centering
\caption{The impact of scaling the amount of in-context data provided to the model is studied. Generally, increasing the amount of data improves performance, but accuracy improvements after 20 in-context points are marginal. The model's performance is strong even with a very small number of in-context data points (e.g., 5).}
\label{tab:context_scaling}
\begin{tabular}{ccccc}
\toprule
Number of In-Context Observations &
\shortstack{US} &
\shortstack{Alberta} &
\shortstack{Australia} &
\shortstack{UK} \\
\midrule
1 & 6.15 & 3.65 & 7.50 & 6.86 \\
2 & 5.46 & 2.85 & 6.72 & 6.26 \\ 
5 & 4.96 & 2.50 & 6.32 & 5.73 \\
10 & 4.84 & 2.34 & 6.21 & 5.94 \\
20 & 4.75 & 2.22 & 6.17 & 5.39 \\
40 & 4.73 & 2.19 & 6.21 & 5.34 \\
\bottomrule
\end{tabular}
\end{table}

\subsection{Thermal complexity of the Australian out-of-distribution test case}
\label{sec:aus-thermal-complexity}
Predicting subsurface temperature in Australia is challenging because the continent couples an ancient, largely cratonic lithosphere with strong intraplate thermal heterogeneity. 
Regional heat-flow syntheses indicate pronounced thermal anomalies in central and eastern Australia that are not systematically correlated with crustal thickness or simple surface geology, reflecting instead lateral variations in crustal heat production and complex lithospheric processes \cite{neumann2000regional}. 
Such nonuniform heat production and spatially variable boundary conditions violate the assumptions of laterally smooth, stationary conductive models and increase the difficulty of capturing temperature fields with regional empirical predictors.
This challenge is particularly acute in the Cooper–Eromanga system. 
The region comprises multiple vertically stacked sedimentary basins developed above radiogenically enriched Proterozoic basement, resulting in a complex thermal architecture strongly influenced by basement composition and long-term burial history \cite{nixon2024thermal}. 
Basin-scale thermal evolution studies indicate that progressive sedimentary burial and transient tectono-hydrothermal heating, particularly during the mid-Cretaceous, have overprinted earlier thermal regimes and generated pronounced lateral and vertical heterogeneity in present-day subsurface temperatures \cite{roth2022down}. 
As a consequence, temperature in the Cooper–Eromanga system does not follow a smooth or stationary relationship with depth or location, complicating predictive modeling approaches based on steady-state or regionally averaged assumptions.

\subsection{Data distribution}
\begin{figure}[H]
    \centering
    \includegraphics[width=\textwidth]{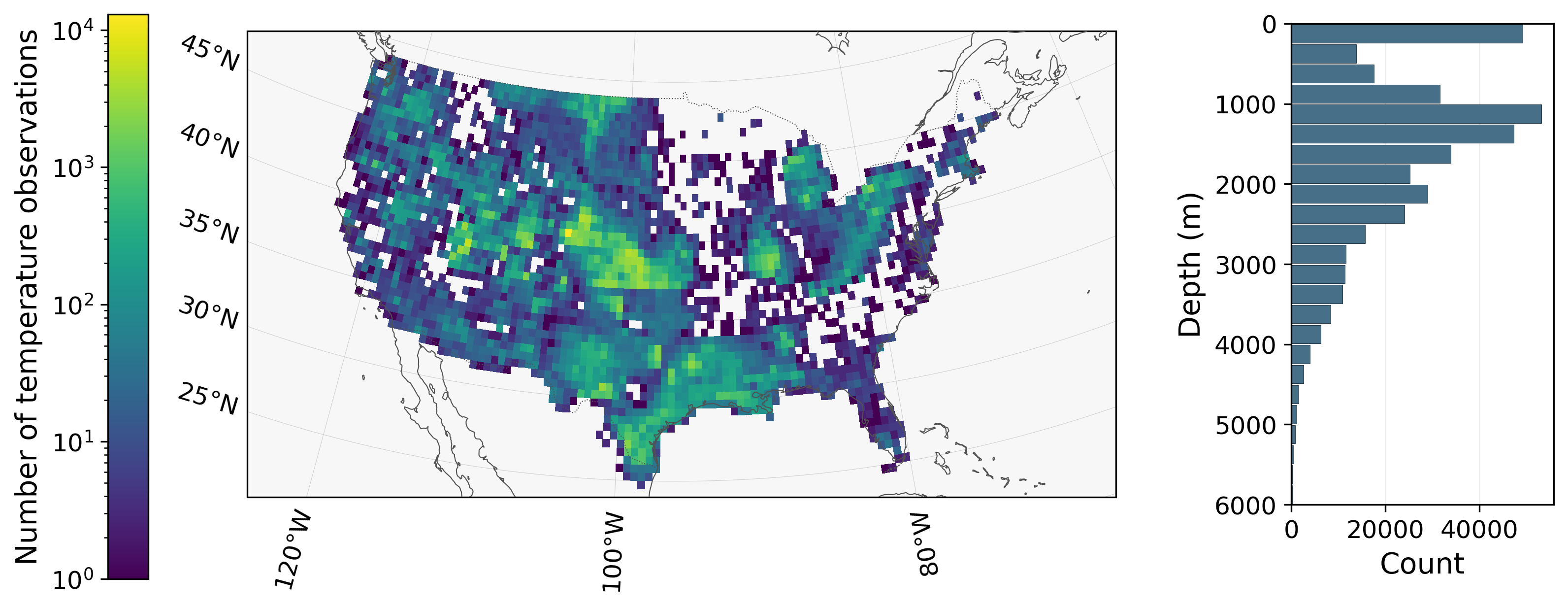}
    \caption{Spatial and depth distribution of bottomhole temperature observations in the contiguous US. The map shows the number of temperature observations in each 0.5$^\circ$ by 0.5$^\circ$ latitude-longitude cell. The histogram shows the number of temperature observations as a function of depth.}
    \label{fig:data_us_temperature_observations}
\end{figure}

\begin{figure}[H]
    \centering
    \includegraphics[width=\textwidth]{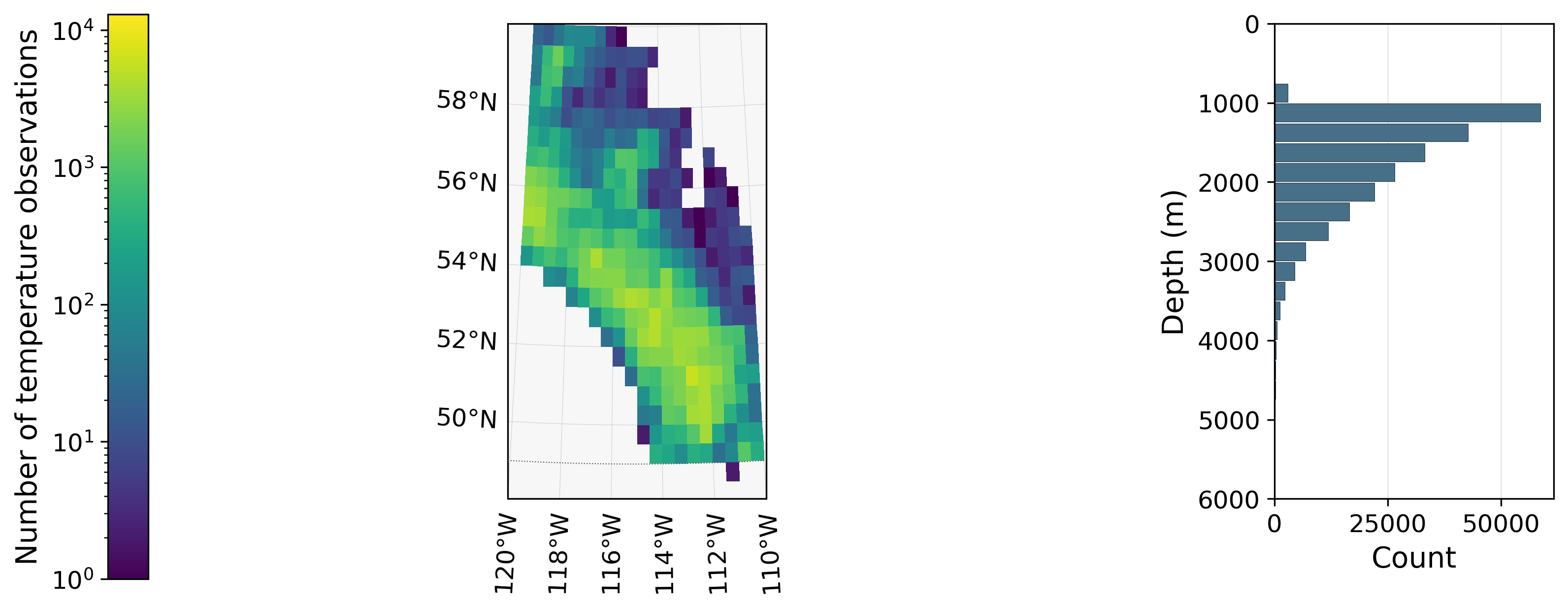}
    \caption{Spatial and depth distribution of bottomhole temperature observations in Alberta, Canada. The map shows the number of temperature observations in each 0.5$^\circ$ by 0.5$^\circ$ latitude-longitude cell. The histogram shows the number of temperature observations as a function of depth.}
    \label{fig:data_alberta_temperature_observations}
\end{figure}

\begin{figure}[H]
    \centering
    \includegraphics[width=\textwidth]{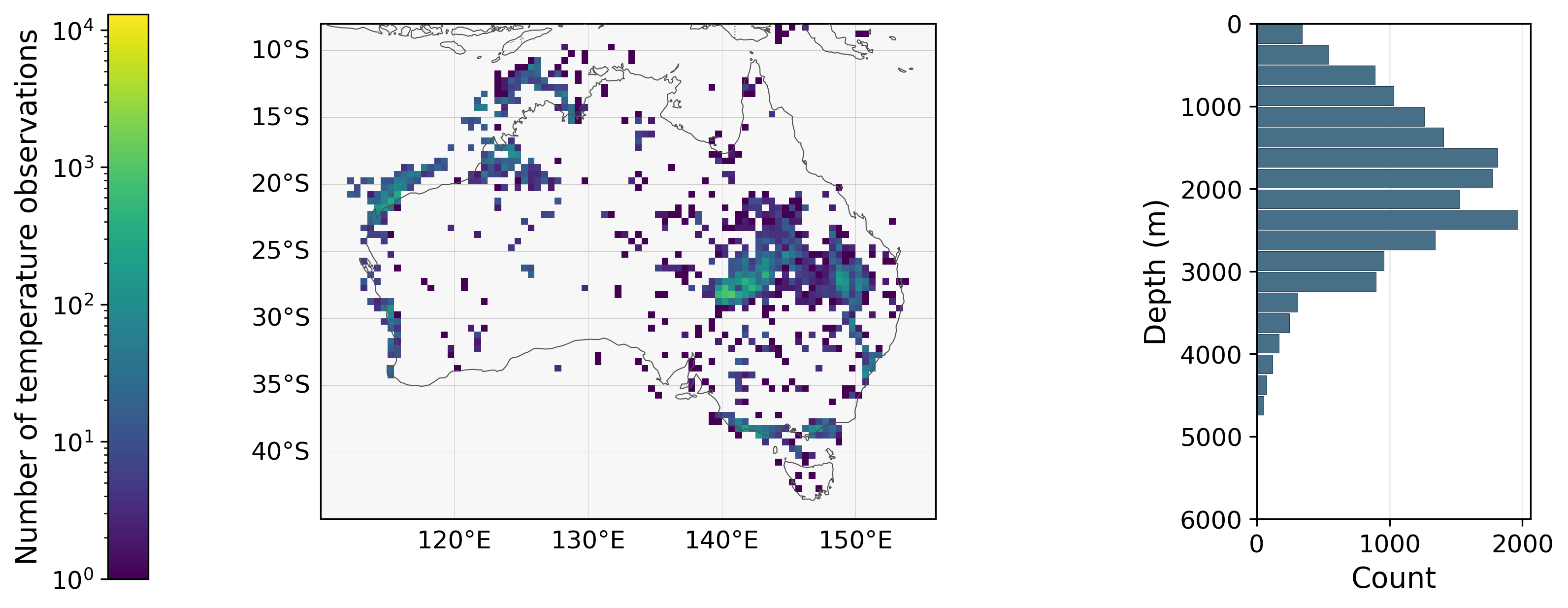}
    \caption{Spatial and depth distribution of bottomhole temperature observations in Australia. The map shows the number of temperature observations in each 0.5$^\circ$ by 0.5$^\circ$ latitude-longitude cell. The histogram shows the number of temperature observations as a function of depth.}
    \label{fig:data_australia_temperature_observations}
\end{figure}

\begin{figure}[H]
    \centering
    \includegraphics[width=\textwidth]{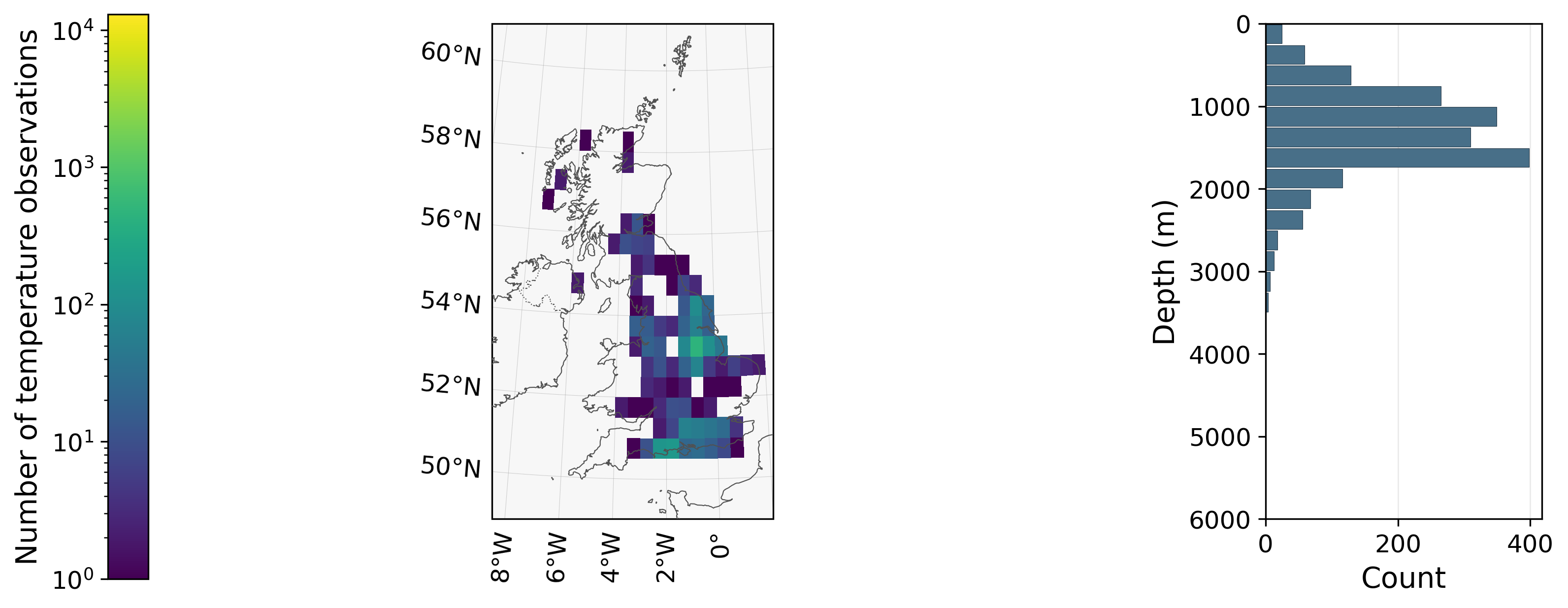}
    \caption{Spatial and depth distribution of bottomhole temperature observations in the UK. The map shows the number of temperature observations in each 0.5$^\circ$ by 0.5$^\circ$ latitude-longitude cell. The histogram shows the number of temperature observations as a function of depth.}
    \label{fig:data_uk_temperature_observations}
\end{figure}

\subsection{Uncertainty versus distance to in-context data}
\begin{figure}[H]
    \centering
    \includegraphics[width=1\linewidth]{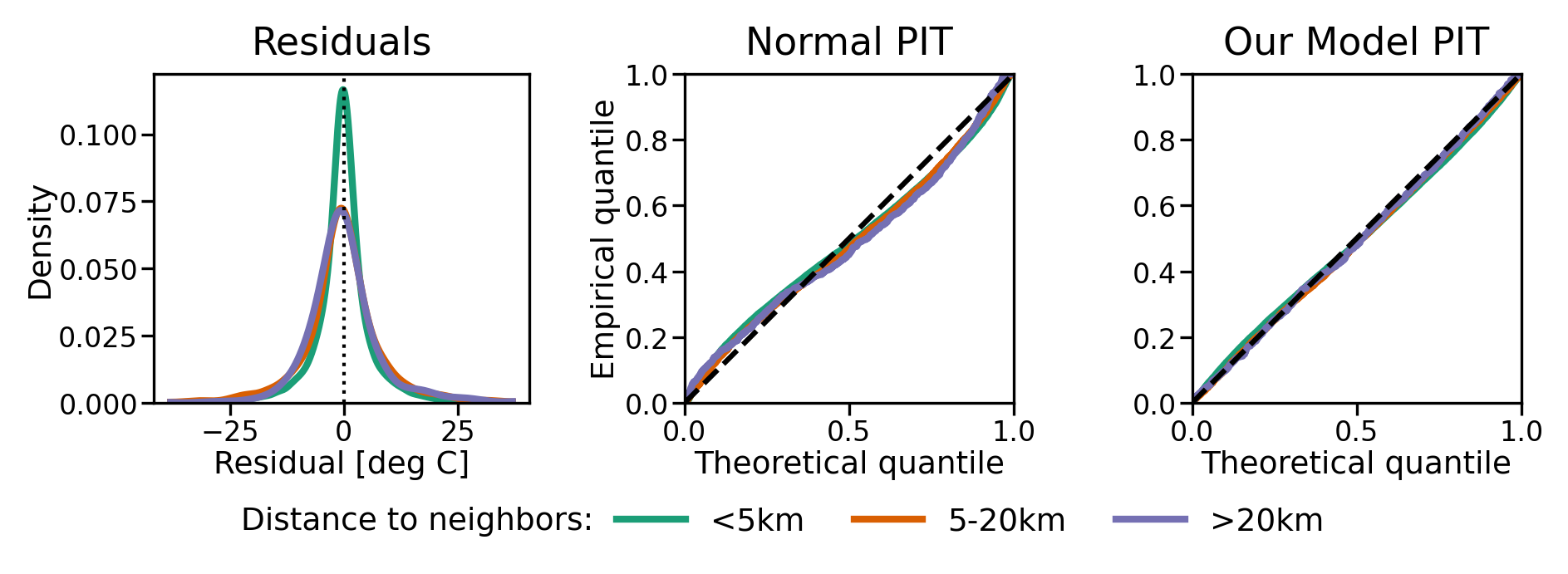}
    \caption{(Left) The distribution of residuals for temperature predictions as a function of the average distance to the in-context data with green being 0-5 km, orange being 5-20 km, and purple being 20 km or further. (Center) and (Right) show the Q-Q plots for the normal approximation to these distributions and our model. Note that our model remains closer to the dashed line for all distances, similar to the version shown in the main text as a function of depth.}
    \label{fig:error_vs_distance}
\end{figure}

\subsection{Alternative training and validation splits}
Because In-Context Earth makes predictions by conditioning on nearby measurements, the usual distinction between training and validation data is more subtle than in standard supervised machine learning.
To address this, we evaluated alternative training and validation splits to test two related concerns: whether performance depends on reusing information from the same borehole, and whether the model remains accurate when validation targets are drawn from spatially held-out regions but local observations are still available in context.
We combined the bottomhole temperature measurements from all four countries into a single dataset and constructed three 90/10 train/validation partitions from the combined data.
One is a random split where individual measurements were assigned independently to the training or validation set, so measurements from the same borehole could appear in both sets in rare cases where multiple temperatures from the same borehole are in the dataset.
This is analogous to the split used in the main text, but using data from all four countries for training instead of just data from the US.
In the borehole-disjoint split, all measurements sharing the same latitude/longitude coordinate pair were assigned to the same partition, ensuring that no borehole location appeared in both training and validation data.
In this case, only data from the training set are provided in context during validation, so the score reflects a scenario where the data from the target borehole are never provided in context.
In the spatial-block split, measurements were grouped into $1^\circ\times1^\circ$ latitude/longitude cells using integer-degree spatial blocks, and each block was assigned wholly to either training or validation data.
For the spatial-block split, in-context observations for validation predictions were drawn from the validation split excluding the target measurement, so this evaluation measures few-shot prediction within held-out spatial blocks rather than predictions without local observations, similar to the held-out continents in the main text.
All three splits used approximately 90\% of measurements for training and 10\% for validation.
For all three evaluations, validation targets were excluded from the in-context set used to make their predictions.
All three splits provide strong performance, comparable to or better than the performance of the model in the main text (note, the training dataset size is larger than in the main text).
These models are all trained using the configuration in the rightmost column of Table \ref{tab:ablation-training-details}.

\begin{table}[H]
\centering
\caption{Validation MAEs for three different training splits that reflect random splitting, ensuring target borehole data are never provided in context, and a block-based split. All splits combine data from all four countries studied in the main text and use a 90/10 split between training and validation.}
\label{tab:mae_other_splits}
\begin{tabular}{lc}
\toprule
Split & MAE $^\circ$C \\
\midrule
Random & 3.9 \\
Borehole disjoint & 4.2 \\
1$^\circ$ block & 4.9 \\
\bottomrule
\end{tabular}
\end{table}

\subsection{Analysis of extreme residual outliers in the test data}

We analyzed the largest differences between the model's predictions in each of the three countries used for testing (Canada, the UK, and Australia).
In the case of the UK and Australia there were data quality issues.
In Canada there were no extremely large errors, and the worst outlier is not far outside the model's 95\% confidence interval.
See Table \ref{tab:bht_largest_outliers} for more details on the locations, temperatures, and predictions.

In the UK, the largest error comes from the data point that has the highest recorded temperature in the UK dataset and is an ``MWD Recording,'' which is present in 10\% of UK data.
This indicates the temperature was measured while drilling and may not be reflective of the formation temperature.
In the same well at similar depths (within 20 meters), there are other measurements that range from 66 to 150 $^\circ$C.
All but one of these are also measured while drilling so potentially unreliable.
The one that is measured with the more reliable wireline logging shows a temperature of 73 $^\circ$C at a depth of 1725 meters, which is well-aligned with the model's prediction.
The second largest residual in the UK is the 150 $^\circ$ temperature observation in the same well.

In Australia, the measurement associated with the most extreme residual is flagged with ``temperature may be assigned,'' which is present in about 147 temperature observations in Australia (almost 1\% of the data).
Similar to the UK, there is other nearby temperature data within 20 meters that has a wide range of temperature values (105 $^\circ$C to 220 $^\circ$C).
An additional consideration in Australia is that this outlier is the second hottest temperature in Australia despite being in the Canning basin, whereas other hot spots in Australia are all in the Cooper-Eromanga basin.
The next hottest data point in the Canning basin is more than 75 $^\circ$C cooler, despite being more than 1700 meters deeper.
The second largest outlier in Australia is marked as ``Error in depth temp'' and the depth is set to 7420 meters, so the model cannot accurately predict this.
The ``Error in depth temp'' flag appears in only 16 data points.

This analysis indicates that extreme discrepancies between the model's predictions and the test data, such as those in the UK and Australia, are due to data quality issues more so than modeling errors.
We chose not to filter out such observations from our test data to avoid a confirmation bias that would result from selectively examining data where the model's predictions do not match the data.

\begin{table}[H]
\centering
\caption{Largest bottomhole temperature residual outliers are shown in each of the three out-of-distribution test regions.}
\label{tab:bht_largest_outliers}
\begin{tabular}{lrrrrrlp{4.5cm}}
\toprule
Region & Lat. & Lon. & \shortstack{Depth\\(m)} & \shortstack{Temp.\\($^\circ$C)} & \shortstack{Pred.\\temp.\\($^\circ$C)} & \shortstack{95\% CI\\($^\circ$C)} & Data quality concerns \\
\midrule
UK & 50.683 & -1.983 & 1590.5 & 155.0 & 72.0 & [55.1, 90.8] & Measured while drilling, hottest temperature in the UK, much cooler observations nearby \\
Australia & -19.247 & 122.346 & 2391.0 & 242.0 & 126.5 & [81.2, 196.8] & ``temperature may be assigned'', much cooler observations nearby, second-highest temperature in Australia despite being in the Canning basin whereas other Australian hot spots are all in the Cooper-Eromanga basin \\
Alberta & 54.288 & -115.661 & 2423.1 & 87.0 & 46.0 & [23.9, 83.0] & None \\
\bottomrule
\end{tabular}
\end{table}

\subsection{Comparison of subsurface representations}
\begin{figure}[H]
    \centering
    \includegraphics[width=1\linewidth]{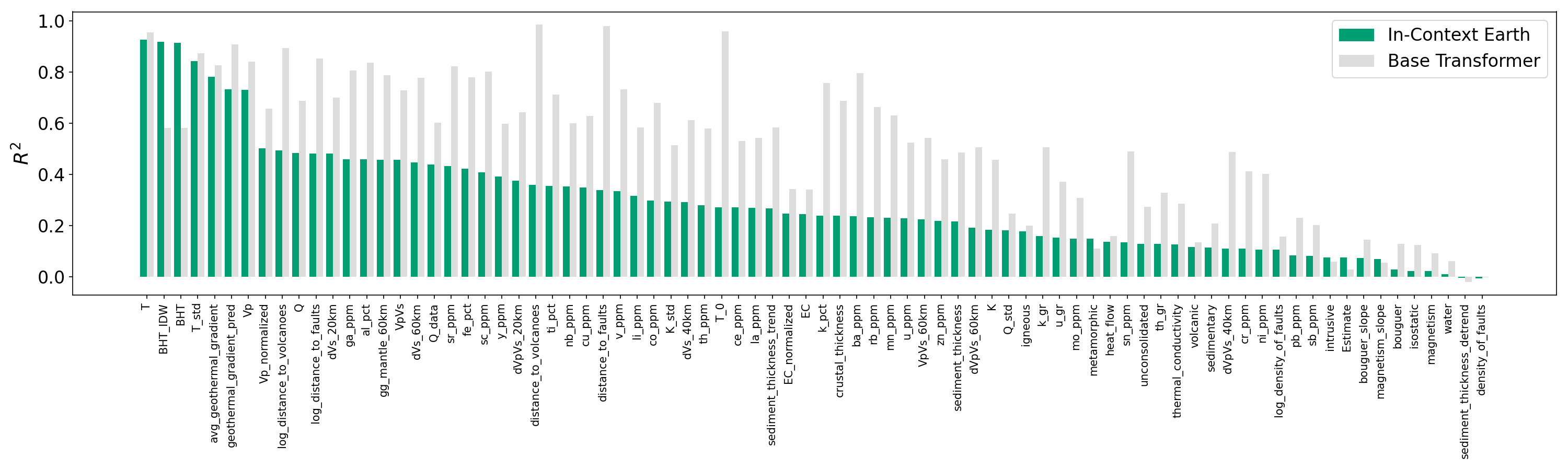}
    \caption{A comparison between the representations of various subsurface properties produced by a base transformer model and In-Context Earth. The base transformer model often learns representations that have higher $R^2$ values but frequently fails to use these representations in a physically sensible way. In-Context Earth has the opposite behavior.
    }
    \label{fig:representation_comparison}
\end{figure}

\subsection{Nuisance-controlled representation probing}
\begin{figure}[H]
    \centering
    \includegraphics[width=0.5\linewidth]{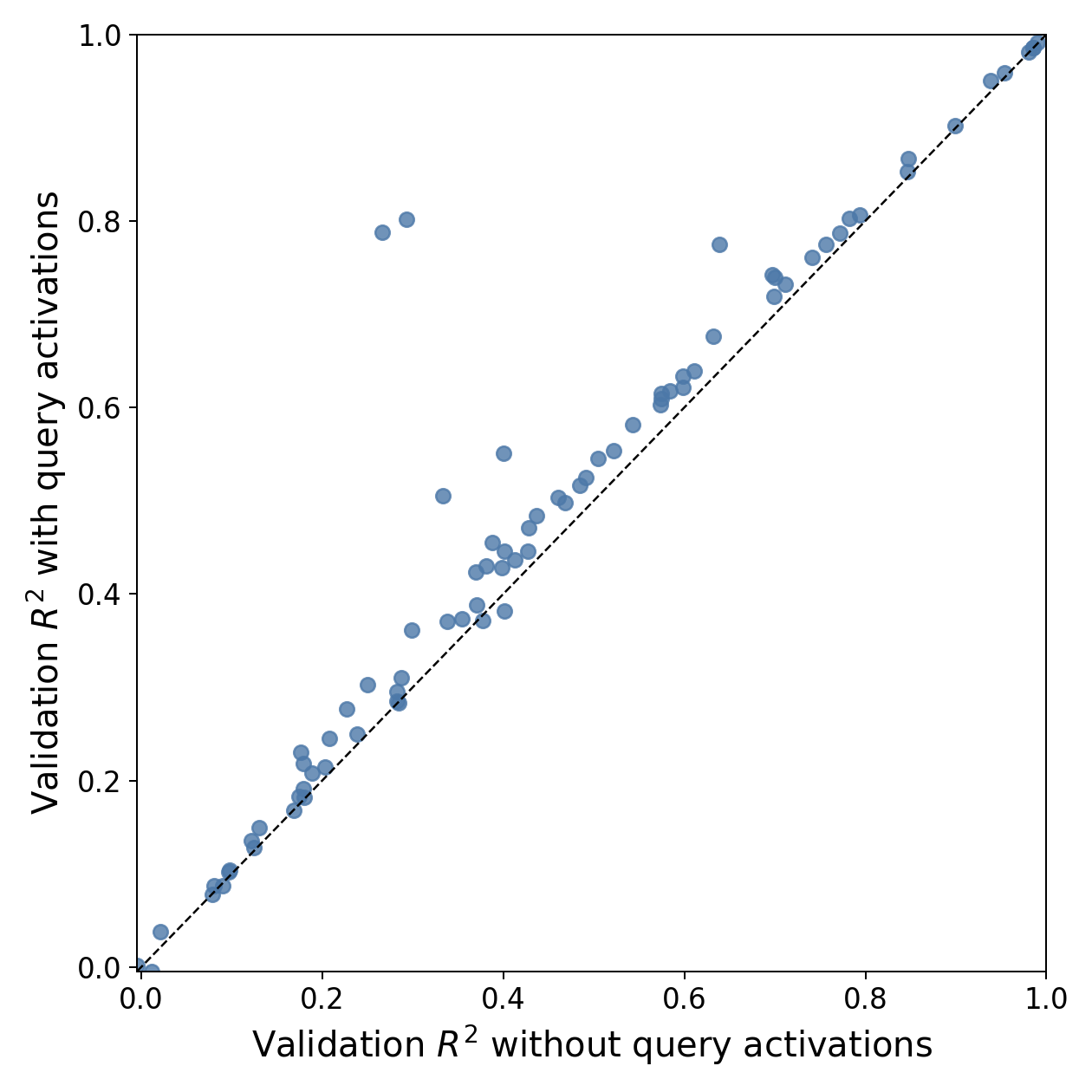}
    \caption{Nuisance-controlled representation probing. Each point compares held-out $R^2$ from a nuisance-only ridge probe with a probe that also includes the final-layer query activation. Adding the query activation increases $R^2$ for 73 of 80 Stanford Thermal Model variables. Points above the dashed line indicate variables for which the query activation adds predictive information beyond the nuisance covariates.
    }
    \label{fig:nuisance_probe}
\end{figure}

To test whether the probed Stanford Thermal Model variables were explained by
simple nuisance covariates, we performed a nested ridge-probing comparison
(Figure \ref{fig:nuisance_probe}). For each variable, we first fit a
ridge model using only nuisance covariates, including query position (latitude, longitude, depth, and elevation), in-context temperature summaries (mean, standard deviation, minimum, maximum, median, closest-neighbor temperature, inverse-distance-weighted mean, and local temperature-depth gradient), in-context distance statistics (minimum, mean, and maximum distance to the neighboring temperature observations), relative-depth statistics (the mean and standard deviation of neighbor depth minus query depth), and the model-predicted query temperature. We then fit a second ridge model that also included the final-layer query activation and measured the incremental $R^2$ relative to the nuisance-only model.

The addition of the query activation improved held-out $R^2$ for 73 of the 80 Stanford Thermal Model variables and decreased it for only 7, indicating that the vast majority of probed variables are better predicted when the model's internal representation is included. In Figure \ref{fig:nuisance_probe}, this appears as most points lying above the dashed line. Positive incremental $R^2$ values indicate that the final-layer representation contains linearly decodable information about these variables beyond what is captured by this nuisance covariate set.

\end{document}